\documentclass[times,twocolumn,preprint,3p,authoryear]{elsarticle}

\usepackage{ycviu}
\usepackage{framed,multirow}

\usepackage{amssymb}
\usepackage{latexsym}

\usepackage{url}
\usepackage{xcolor}
\definecolor{newcolor}{rgb}{.8,.349,.1}

\usepackage{subfiles}
\usepackage[caption=false]{subfig}
\usepackage[export]{adjustbox}
\usepackage{booktabs}
\usepackage{xcolor}
\usepackage{boldline}
\usepackage{ulem}
\usepackage[ruled,lined,boxed]{algorithm2e}
\usepackage{nccmath}
\usepackage{makecell, tabularx}
\usepackage{rotating}
\usepackage{comment}
\usepackage{enumitem}
\usepackage{pdfpages}

\newcommand{\redul}[1]{{\color{red}\uline{{\color{black}#1}}}}
\newcommand{\blueul}[1]{{\color{blue}\uline{{\color{black}#1}}}}

\DeclareMathOperator*{\argmin}{arg\,min}

\journal{Computer Vision and Image Understanding}

\begin{document}

\ifpreprint
  \setcounter{page}{1}
\else
  \setcounter{page}{1}
\fi

\begin{frontmatter}

\title{Exploiting Image Translations via Ensemble Self-Supervised Learning for Unsupervised Domain Adaptation}%


\author[1]{\corref{cor1}Fabrizio J. \snm{Piva}}
\ead{f.j.piva@tue.nl}
\author[1]{Gijs \snm{Dubbelman}}

\cortext[cor1]{Manuscript under review at Computer Vision and Image Understanding}

\address[1]{Eindhoven University of Technology, Department of Electrical Engineering, Groene Loper 12, 5612AZ Eindhoven, The Netherlands}

\received{1 May 2013}
\finalform{10 May 2013}
\accepted{13 May 2013}
\availableonline{15 May 2013}
\communicated{S. Sarkar}

\begin{abstract}
    We introduce an unsupervised domain adaption (UDA) strategy that combines multiple image translations, ensemble learning and self-supervised learning in one coherent approach. We focus on one of the standard tasks of UDA in which a semantic segmentation model is trained on labeled synthetic data together with unlabeled real-world data, aiming to perform well on the latter. To exploit the advantage of using multiple image translations, we propose an ensemble learning approach, where three classifiers calculate their prediction by taking as input features of different image translations, making each classifier learn independently, with the purpose of combining their outputs by sparse Multinomial Logistic Regression. This regression layer known as meta-learner helps to reduce the bias during pseudo label generation when performing self-supervised learning and improves the generalizability of the model by taking into consideration the contribution of each classifier. We evaluate our method on the standard UDA benchmarks, i.e. adapting GTA V and Synthia to Cityscapes, and achieve state-of-the-art results in the mean intersection over union metric. Extensive ablation experiments are reported to highlight the advantageous properties of our proposed UDA strategy.
\end{abstract}%

\end{frontmatter}



\section{Introduction}
\label{sec:intro}

Recently, deep learning has shown impressive results in many computer vision tasks. This advancement has largely come as a result of training very deep neural networks on large-scale datasets~\citep{imagenet2009cvpr, openimages}. The satisfactory performance of these models is substantially tied to the training data due to dataset bias~\citep{datasetbias}, and these models are unfortunately incapable of generalizing well to unseen data. To circumvent this limited generalization of deep models to unseen data, the goal of Unsupervised Domain Adaptation (UDA) is to improve the model's performance on an \textit{a priori} known target dataset without requiring that this target dataset is labeled. 

\begin{figure}[t!]
\centering
\subfloat[][Multi-task tri-training\\~\citep{Ruder2018StrongBF, Zhang2018ICASSP}]{\label{fig:mtri}
    \includegraphics[height=0.12\textheight]{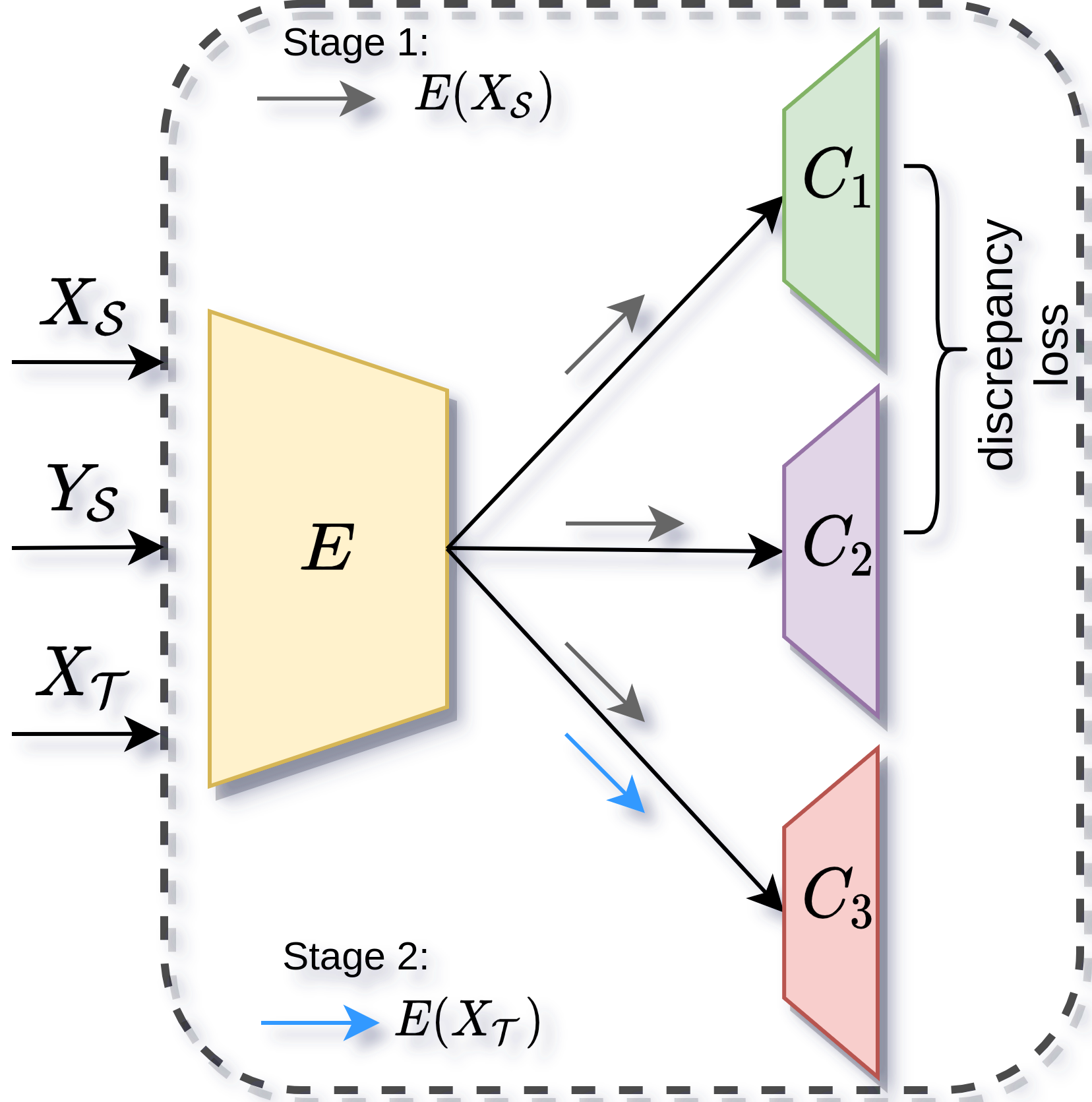}
    }\qquad
\subfloat[][Our approach]{\label{fig:contribution}
\includegraphics[height=0.12\textheight]{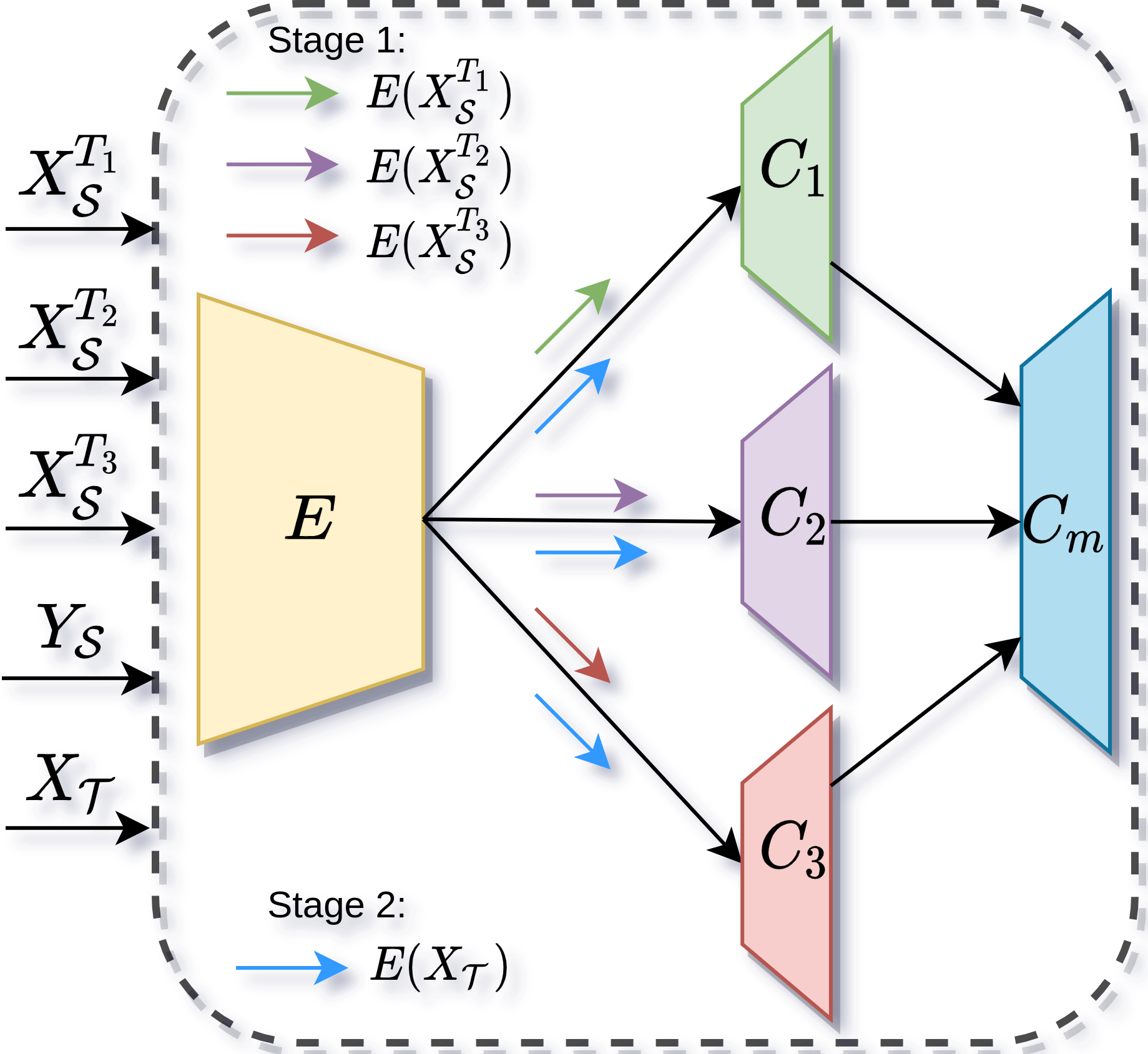}}
\caption{\label{fig:eyecatch}To adapt a source annotated set to an unlabeled target set,~\citep{Ruder2018StrongBF, Zhang2018ICASSP} leverages the images and labels of the source set to train three classifiers $C_1$,~$C_2$, and $C_3$ in a supervised way, enforcing discrepancy between $C_1$ and $C_2$ via a cosine distance loss. Once trained, only the predictions of $C_1$ and $C_2$ are used to label the target set, and $C_3$ is trained separately on this new labeled dataset, obtaining the final inference from $C_3$. Our proposed ensemble uses three image translations of the source dataset, and encourages discrepancy across all classifiers by feeding the features of each translation to a different predictor. Hereafter, the meta-learner $C_m$ learns to weight the predictions of each classifier for the classes, to create pseudo-labels for the target set. Unlike~\citep{Zhang2018ICASSP}, a round of self-training will consist of training $C_1$, $C_2$ and $C_3$ on this new labeled target set, and retraining $C_m$ to create robust pseudo-labels. Finally, inference is made via the meta-learner $C_m$.} 
\end{figure}

In UDA the model's training data contains both fully labeled samples from a source domain as well as unlabeled data from the target domain. In general, models that are conventionally trained with data from two (source and target) distributions suffer a significant performance drop due to the underlying domain gap. For this reason, UDA approaches aim to mitigate this gap by transferring the knowledge learned from the source annotated domain to the unlabeled target domain. Although the nature of the source and target distributions can vary depending on the application and computer vision task, one of the most challenging scenarios involves training a semantic segmentation model using synthetic data as the (labeled) source dataset and real-world data as the (unlabeled) target dataset. Since this synthetic-to-real per-pixel classification scenario involves a domain gap that is very challenging to address, it is used as the standard scenario in practically all recent computer vision research on UDA \citep{Li2019BDLCVPR, Luo2018cvpr, yang2020fda, yang2020labeldriven, Zou2018eccv}. To be able to compare our work with these state-of-the-art methods using the standard UDA benchmarks, our work also focuses on performing UDA in the context of synthetic-to-real training of semantic segmentation models. However, we emphasize that the applicability and practical value of UDA are broader than this particular scenario.

Lately, several UDA methods have shown promising results by using self-supervised learning (SSL), a technique that leverages the model's predictions to label the target domain and retrain the model on this new subset. To determine whether a prediction is a label candidate or not, a criterion needs to be established, and current single encoder-decoder methods adopt a confidence thresholding scheme as the standard procedure~\citep{Li2019BDLCVPR, yang2020fda, Zou2018eccv}. A deficiency of this thresholding approach is that the model can still consider high confident mistaken predictions as pseudo-labels, affecting negatively the retraining process. To address this issue, \citep{Ruder2018StrongBF} investigate several ensemble approaches that leverage multiple classifiers. With this adjustment, an extra condition can be added to complement confidence thresholding: if the classifiers agree on the winner class, the prediction can be considered as pseudo-label.

In the particular ensemble approach Multi-task tri-training proposed in~\citep{Ruder2018StrongBF}, one encoder is shared across three classifiers and the training is performed in two stages (see Fig.~\ref{fig:mtri}). First, the source annotated set is used to train the encoder along with all three classifiers in a supervised manner, while disagreement between the first two classifiers is enforced with a discrepancy loss, defined as the cosine distance between the weights of the first two classifiers~\citep{Bousmalis2016}. In the second training stage, these two classifiers will create pseudo-labels for the unlabeled target set using both confidence thresholding and class agreement as labeling criteria, and the third classifier will be trained on this labeled subset of the target set. This process is repeated a certain number of times (known as self-supervision rounds) until convergence.

Although this general approach of \citep{Ruder2018StrongBF} is suitable for UDA, its practical implementation for semantic segmentation~\citep{Zhang2018ICASSP} has shown limited performance. This can be attributed to several pitfalls in the training strategy: 1) the model does not fully exploit all the members of the ensemble for pseudo label generation, since it uses only two out of the three classifiers for this process, and 2) the cosine distance between the weights of two classifiers to encourage discrepancy implies that the angle between the weights of two classifiers will converge to $90$ degrees, but this is not a sufficient condition to ensure useful disagreement between the two classifiers. 

To overcome the aforementioned deficiencies, we research an alternative approach in which all the classifiers participate during pseudo-label generation as well as network retraining, and where the discrepancy is encouraged without a cosine distance loss (see Fig.~\ref{fig:contribution}). We hypothesize that a discrepancy loss is not needed if instead, each classifier learns from a different set of features. Given a source annotated image, we propose in the first training stage to leverage multiple different image-to-image translations, making sure that each classifier along with the encoder learns from a particular translation. As a result, each predictor will focus on a particular translation and therefore we eliminate the need of a specific (cosine) loss for disagreement. In addition, our second contribution is a meta-learning layer that ensembles the output of each classifier for the classes, considering a classifier more than the others when it performs better than the rest for a particular class. During the second stage of SSL, the meta-learner is trained on the outputs of the three classifiers to generate robust pseudo-labels on the target set that are used to retrain the entire network. This process is repeated for few iterations until convergence is reached.

Besides self-supervised learning, practically all state-of-the-art UDA methods, of which the most relevant ones are detailed in Section \ref{sec:mtrivssota}, exploit a combination of techniques that target different aspects of the UDA problem. The most commonly used techniques are:

\begin{itemize}[noitemsep,nolistsep]
    \item Image translations: creating alternative representations of either the source, the target, or both domains to reduce the domain gap between source and target and thereby increase the robustness of the model~\citep{Gong_2019, Murez18, DCAN2018, yang2020labeldriven,  yang2020fda}.
    \item Feature alignments: using adversarial training to align the features from both domains without supervision to reduce the domain gap at feature level~\citep{Hong2018CVPR, Romijnders19WACV, Swami2017UDA, Tsai2018CVPR, Yiheng2018CCPR}.
    \item Model regularization: preventing the model from overfitting to the most dominant classes by regulating the probability distribution on the output space~\citep{TH2018ADVENTCVPR, yang2020fda}.
\end{itemize}
Although the focus of this research is to improve self-supervised learning in the context of UDA through combining ensemble learning with multiple image translations, we also integrate the aforementioned techniques in our approach, to reach state-of-the-art performance. The details of our approach are provided in Section \ref{sec:method}. In our experiments, described in Section \ref{sec:experiments}, we take care to differentiate between the performance obtained using the complete set of techniques and the performance obtained as a result of our novel methodology.

In summary, the main contributions of our work are:

\begin{itemize}[noitemsep,nolistsep]
    \item A meta-learner that exploits multiple image-to-image translations within the context of Ensemble Learning via weighting each classifier's prediction with a sparse Multinomial Logistic Regression. This opens a new line of research where Ensemble Learning~\citep{ensemble92} and Multitask tri-training~\citep{Ruder2018StrongBF} meet.
    \item Our approach achieves state-of-the-art performance for two standard UDA benchmarks: adapting GTA V~\citep{Richter2016ECCV} to Cityscapes~\citep{Cordts2016Cityscapes} and SYNTHIA~\citep{synthia} to Cityscapes.
\end{itemize}

\begin{figure*}[htb]
 \centering
 \includegraphics[height=0.1447\textheight]{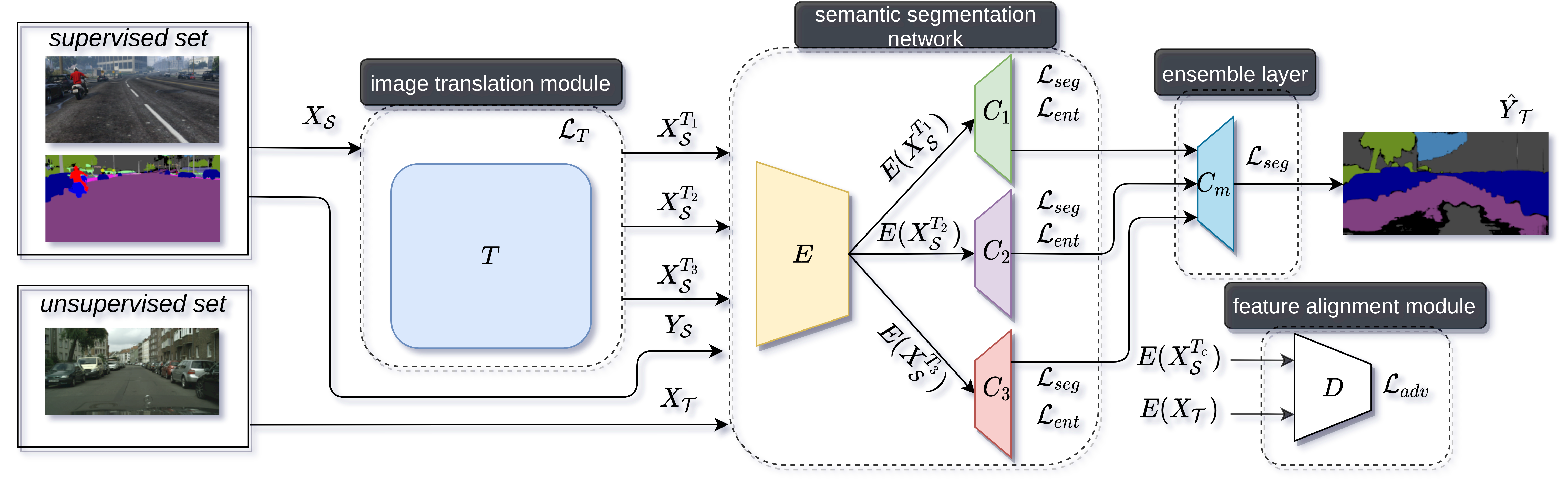}
 \caption{\textit{Network architecture and losses}. Our method combines different transformations coming from an image translation module $T$ by teaching a meta-learner $C_m$ to balance the outputs from the classifiers $C_1$, $C_2$ and $C_3$ with respect to the classes. Each classifier focuses on the features of a single transformation, and the feature extractor $E$ acts as a common features provider. Extra adaption is encouraged through $D$ via adversarial learning to align the features of both source and target distributions.}
 \label{fig:proposedmethod}
\end{figure*}

\section{Related work}

Most of the state-of-the-art methods combine different strategies to achieve competitive results. In this section, we focus on related work that, similar to our approach, use strategies such as image translation, self-supervised learning, and/or ensemble methods.

\textbf{Image translation methods for UDA} have recently been widely used to improve the performance of UDA methods. Since these techniques can be trained without the need of labels, the images of the source dataset can be transformed to a different space where the domain gap between the transformed images and the target dataset is smaller. This is the case for several UDA methods, that choose to transform the source to the target set, either by using a deep neural network~\citep{Hoffman18ICML, Li2019BDLCVPR, DCAN2018} or image processing techniques such as the Fourier transform~\citep{yang2020fda}. Other works have considered mapping to an intermediate space~\citep{Murez18}, where the features are domain agnostic. In addition, \citep{yang2020labeldriven} has shown that mapping the target dataset to the source is also effective, leading to state-of-the-art results. Alternatively, \citep{Gong_2019} explores the possibility of generating multiple intermediate representations between the source and the target domain, where each arbitrary representation belongs to a point in a manifold of domains. Regardless of the chosen target space to which the annotated dataset is mapped, to the best of our knowledge, no works have considered using multiple representations in parallel to improve UDA models, as explored in this work.

\textbf{Self-supervised Learning in UDA}. Many recent UDA methods leverage self-supervised learning as a way of using the model's predictions to learn from the unlabeled target domain. When using the model's outputs, it is needed to establish criteria to filter out spurious predictions and select reliable label candidates. Many methods that use a single encoder-single decoder architecture propose as criteria to use confidence thresholding on the probability maps in the output space~\citep{Li2019BDLCVPR, yang2020fda, Zou2018eccv}, although they still suffer from the propagation of errors due to the inclusion of highly confident but mistaken predictions. Other methods such as~\citep{DengLZ19, Xie2020CVPR} prefer to use a teacher-student arrangement, in which the teacher network learns first from both domains to consequently transfer this knowledge to the student model via knowledge distillation~\citep{Nguyen-MeidineB21}. While knowledge distillation has shown promising results, its application to self-supervised learning is limited due to the usage of a single network to generate pseudo-labels, instead of using multiple predictions to agree on the selection of pseudo labels.

\textbf{Ensemble methods for UDA} propose to increase the number of predictions for a single input by changing the network architecture. For instance, Co-Training (CT) utilizes two classifiers to create different points of view from the same sample to produce pseudo-labels~\citep{Blum1998CLU, AH2008}, used later for the unlabeled training data. Recent applications of CT in UDA have demonstrated promising results~\citep{Luo2018cvpr}. Tri-Training (TT)~\citep{ZhiHua2005ieee} can be conceived as an extension of CT, where three members participate in the ensemble, each of these consisting of a feature extractor and a classifier. Since TT is computationally expensive, Multitask tri-training (MTri)~\citep{Ruder2018StrongBF} was proposed, where a common feature extractor is shared among three classifiers, computing different outputs from the same features. The idea behind MTri is to make the feature extractor learn those features that are invariant across the source and target domain, whilst forcing a discrepancy between the classifiers through a discrepancy operator. 

When sharing a feature extractor in MTri, the discrepancy across classifiers becomes a key factor to generate pseudo-labels during SSL~\citep{Zhang2018ICASSP}. While a cosine distance might help to enforce a certain diversity, we hypothesize that feeding constantly the same features to all the classifiers does not optimally allow the encoder to learn domain invariant representations. If we obtain these alternative representations from an image translation model, we can encourage discrepancy by feeding the features of a specific representation to a different classifier, improving simultaneously the generalization capacity of the encoder. These are the principles on which our method is based, bringing together different research lines: image transformations, ensemble learning, and self-supervised learning.

\section{Method}
\label{sec:method}

\subsection{Problem statement}

Given the source dataset $\mathcal{S}$ consisting of a set of images $X_{\mathcal{S}}$ and corresponding semantic labels $Y_{\mathcal{S}}$ (e.g., synthetic data generated by computer graphic simulations) and the unlabeled target dataset $\mathcal{T}$ consisting only of the images $X_{\mathcal{T}}$ without labels, the goal of UDA is to design and train a neural network for semantic segmentation and to make it perform as close as possible to a model hypothetically trained on $X_\mathcal{T}$ with ground truth labels $Y_{\mathcal{T}}$.

\subsection{Network architecture}

The design of our approach is based on the hypothesis that different image-to-image translations of input images can contribute individually to the generalization process. This brings forward a general design, in which an image translation module $T$ generates different image-to-image translations that are consequently used by the semantic segmentation network during UDA training (see Fig.~\ref{fig:proposedmethod}). 

\subsubsection{Image translation module}

Given the high availability of unsupervised image-to-image translation models providing multiple image representations for a single input~\citep{Gatys2016ImageST, huang2017adain, Murez18, Zhu2017Unpaired}, we assume that we can use any of these models for $T$, and therefore we focus our contribution in the ensemble as well as a training strategy. Regardless of the chosen method, $T$ is trained accordingly before starting the first training stage using both sets of images in an unsupervised way $\{X_{\mathcal{S}}, X_{\mathcal{T}}\}$, to obtain the transformations $X^{T_1}_{\mathcal{S}}$, $X^{T_2}_{\mathcal{S}}$ and $X^{T_3}_{\mathcal{S}}$. Implementation details on the used image translation network can be found in Sec.~\ref{sec:impdetails}.

\subsubsection{Semantic segmentation network}

Our proposed semantic segmentation network consists of a shared feature extractor $E$ along with three classifiers $C_1$, $C_2$ and $C_3$. To integrate the transformations $X^{T_1}_{\mathcal{S}}$, $X^{T_2}_{\mathcal{S}}$ and $X^{T_3}_{\mathcal{S}}$ from $T$ as well as the source labels $Y_{\mathcal{S}}$, our ensemble approach has a two-stage training process. 

In the first training stage, the classifiers $C_1$, $C_2$ and $C_3$ accumulate knowledge along with the encoder $E$ by computing their predictions from the features $E(X^{T_1}_{\mathcal{S}})$, $E(X^{T_2}_{\mathcal{S}})$ and $E(X^{T_3}_{\mathcal{S}})$ respectively, learning each one from a different image translation while sharing the same set of label maps $Y_{\mathcal{S}}$.

\subsubsection{Ensemble layer}

After the encoder and classifiers are trained, we need to train the meta-learner to ensemble the predictions of the classifiers. This is done by freezing the weights of the semantic segmentation network while feeding the features of the translated images $E(X^{T_1}_{\mathcal{S}})$, $E(X^{T_2}_{\mathcal{S}})$ and $E(X^{T_3}_{\mathcal{S}})$ into the classifiers $C_1$, $C_2$ and $C_3$ respectively. With these predictions along with the ground truth labels $Y_{\mathcal{S}}$, the meta-learner $C_m$ learns to weigh each classifier's prediction for the classes. The resulting layer $C_m$ condenses rich information as it is capable of balancing each classifier's output to create a single prediction. This is needed to create reliable pseudo-labels for the unlabeled target domain.

The second training stage consists in self-supervised learning, where the predictions of the classifiers on the target images $C_1(E(X_{\mathcal{T}}))$, $C_2(E(X_{\mathcal{T}}))$ and $C_3(E(X_{\mathcal{T}}))$ are used as input for $C_m$, from which the first set of pseudo-labels $\hat{Y}^{(0)}_{\mathcal{T}}$ are obtained using confidence thresholding as in~\citep{Li2019BDLCVPR, yang2020fda}. But our confidence is based on the output of three classifiers combined in an ensemble instead of one classifier, making the pseudo-labels more reliable. By using the pair of target images and pseudo-labels $\{X_{\mathcal{T}}, \hat{Y}^{(0)}_{\mathcal{T}}\}$, the entire semantic segmentation network is retrained, concluding the first training round. For the consecutive rounds of SSL, the meta-learner is retrained using the classifiers' prediction on the target images along with the pseudo-labels of the previous round. After that, the outputs of $C_m$ are used to create new pseudo-labels for the target images to finally retrain the semantic segmentation network.

\subsubsection{Feature alignment module}

During both training stages, the discriminator $D$ is responsible for performing feature alignments with adversarial training between $E(X_{\mathcal{T}})$ and $E(X^{T_c}_{\mathcal{S}})$ where $X^{T_c}_{\mathcal{S}}$ is determined by the image translation module in use. Implementation details on this can be found in Sec.~\ref{sec:impdetails}.

\subsection{Training objectives}

\subsubsection{First training stage}

Given the source annotated set of images $X_\mathcal{S}$ with label maps $Y_\mathcal{S}$, and the alternative sets of image translations $X^{T_1}_{\mathcal{S}}$, $X^{T_2}_{\mathcal{S}}$ and $X^{T_3}_{\mathcal{S}}$, the semantic segmentation outputs from each classifier $C_{k}$, are computed from the features $E(X^{T_k}_{\mathcal{S}})$ of each transformation and used to train the encoder as well as the classifiers during the first training stage: 

\begin{equation}
\label{eq:stage1}
\begin{split}
  \mathcal{L}_{stage1} &=\mathcal{L}_{seg}(C_{k}(E(X^{T_k}_{\mathcal{S}})), Y_\mathcal{S}) +\lambda_{adv}\mathcal{L}_{adv}(X^{T_c}_\mathcal{S}, X_\mathcal{T}) \\
  &+ \lambda_{ent}\mathcal{L}_{ent}(C_{k}(E(X_{\mathcal{T}})))
\end{split}
\end{equation}

$\forall k \in \{1, 2, 3\}$, where the supervised semantic segmentation loss for the classifier $C_k$ is defined by:

\begin{equation}
  \mathcal{L}_{seg}(C_{k}(E(X^{T_k}_{\mathcal{S}})), Y_\mathcal{S})= - \langle Y_\mathcal{S}, \log C_{k}(E(X^{T_k}_{\mathcal{S}}))\rangle 
\end{equation}

The parameter $\lambda_{adv}$ denotes the hyperparameter that controls the relative importance of the adversarial loss. This adversarial component ensures convergence between the features from the transformed images $X^{T_c}_\mathcal{S}$ and the target images $X_\mathcal{T}$ and is defined as:

\begin{equation}
\label{eq:advtraining}
  \mathcal{L}_{adv}(X^{T_c}_\mathcal{S}, X_\mathcal{T})= \mathbb{E}[ \log (D(X^{T_c}_\mathcal{S}))] + \mathbb{E}[ \log(1 - D(X_\mathcal{T})) ]
\end{equation}

where $\mathbb{E}$ represents the expected value operator. 

Considering that entropy minimization has recently shown to improve SSL by means of model regularization~\citep{TH2018ADVENTCVPR, yang2020fda}, $\lambda_{ent}$ represents a scalar that adjusts the weight of the entropy minimization loss $\mathcal{L}_{ent}$. Given the unlabeled target set of images $X_\mathcal{T}$, the classifiers $C_k$ will first compute their predictions $C_k(E(X_\mathcal{T}))$ to regularize their Shannon Entropy~\citep{shannon1948mathematical} with the function: 

\begin{equation}
\label{eq:entropymin}
  \mathcal{L}_{ent}(C_{k}(E(X_{\mathcal{T}})))= \alpha \phi(-\langle C_{k}(E(X_{\mathcal{T}})), \log (C_{k}(E(X_{\mathcal{T}}))) \rangle)
\end{equation}

$\forall k \in \{1, 2, 3\}$, where $\alpha = \frac{-1}{\log(C)}$, and $\phi(x) = (x^2 + 0.001^2)^{\eta}$ is the Charbonnier penality function proposed in~\citep{yang2020fda}, that penalizes high entropy predictions more than the low entropy ones when $\eta > 0.5$, preventing the model from overfitting on the most predominant classes whilst assisting those less present in the dataset. 

To finish the first training stage, the meta-learner $C_m$ is trained to ensemble the outputs of the classifiers $C_1$, $C_2$ and $C_3$ with respect to the transformations of the source dataset. This is done by freezing the semantic segmentation network to obtain the sparse Multinomial Logistic Regression weight vectors $\mathrm{w}_1$, $\mathrm{w}_2$ and $\mathrm{w}_3$ by minimizing the cross-entropy loss: 

\begin{equation}
\label{eq:arbiter_optimization}
\argmin_{\mathrm{w}_1, \mathrm{w}_2, \mathrm{w}_3} \mathcal{L}_{seg}(C_m(X^{T_1}_{\mathcal{S}}, X^{T_2}_{\mathcal{S}}, X^{T_3}_{\mathcal{S}}), Y_\mathcal{S}),
\end{equation}

where the output of the meta-learner is computed for every pixel $(h,w)$ as follows:

\begin{equation}
\label{eq:arbiter}
\begin{split}
  C_m(X^{T_1}_{\mathcal{S}}, X^{T_2}_{\mathcal{S}}, X^{T_3}_{\mathcal{S}})^{(h,w)} &= 
  \mathrm{w}_1 \odot C_{1}(E(X^{T_1}_{\mathcal{S}}))^{(h,w)} \\ + \mathrm{w}_2 \odot C_{2}(E(X^{T_2}_{\mathcal{S}}))^{(h,w)} +
  &\mathrm{w}_3 \odot C_{3}(E(X^{T_3}_{\mathcal{S}}))^{(h,w)}.
\end{split}
\end{equation}

The dimension of the weight vectors $\mathrm{w}_{k}$ is the same as the total number of classes and $\odot$ denotes element-wise multiplication. Because the output of $C_m$ for a given pixel and class depends only on the three classifier outputs for that specific pixel and class, we refer to it as a sparse version of the standard Multinomial Logistic Regression~\citep{bishop06}.

\subsubsection{Second training stage}

The second stage consists mainly in self-supervised learning, a process in which the meta-learner $C_m$ generates pseudo-labels $\hat{Y}_\mathcal{T}$ for the target images $X_{\mathcal{T}}$ to retrain the semantic segmentation network. Since $C_m$ is already trained, we feed the features of the target images $E(X_{\mathcal{T}})$ in all the classifiers in Eq.~(\ref{eq:arbiter}) and apply global confidence thresholding on the probability maps~\citep{Li2019BDLCVPR, yang2020fda} to obtain the initial pseudo-labels $\hat{Y}^{(0)}_\mathcal{T}$. Hereafter, we start the self-supervised learning rounds $i=\{1, 2, ...\}$, each one consisting of three steps. First, the semantic segmentation network is retrained through the loss: 

\begin{table*}[!htp]
 \centering
\caption{\label{tab:abstudy1}Adaptation from GTA V $\xrightarrow{}$Cityscapes, analyzing different architectures as well as the impact of entropy minimization for the first training stage. We show IoU for each class and total mean IoU. Apart from indicating the best IoU in bold, we make an intra comparison between $C_1$, $C_2$ and $C_3$ against $C_m$ when using entropy minimization (underlined with red), and without the entropy loss (underlined with blue). Note that although entropy minimization helps to close the gap between SED and our method, there is still a remarkable difference specially considering the performance of $C_m$.}
\begin{adjustbox}{width=0.95\textwidth}
    \begin{tabular}{l | ccccccccccccccccccc| c}
    \toprule
    \multicolumn{21}{c}{\textbf{GTA V} $\xrightarrow{}\textbf{Cityscapes}$} \\
    \midrule
     Experiment & \rotatebox{60}{road} & \rotatebox{60}{side.} & \rotatebox{60}{buil.} & \rotatebox{60}{wall} & \rotatebox{60}{fence} &  \rotatebox{60}{pole} & \rotatebox{60}{light} & \rotatebox{60}{sign} & \rotatebox{60}{veget.} & \rotatebox{60}{terr.} & \rotatebox{60}{sky} & \rotatebox{60}{person} & \rotatebox{60}{rider} & \rotatebox{60}{car} & \rotatebox{60}{truck} & \rotatebox{60}{bus} & \rotatebox{60}{train} & \rotatebox{60}{motor} & \rotatebox{60}{bike} & \textbf{mIoU} \\ \midrule
    SED (w/o ent) & 77.72 & 35.69 & 78.91 & \textbf{33.2} & 19.95 & 34.86 & 25.25 & 3.28 & 80.77 & 34.37 & 71.85 & 60.06 & 18.46 & 84.6 & 22.66 & 21.41 & 1.09 & 23.29 & 21.82 & 39.43\\
    SED (w/ ent) & 84.89 & 34.30 & \textbf{82.64} & 31.24 & 18.91 & \textbf{36.78} & 32.18 & 15.20 & 82.33 & 31.89 & 72.78 & 63.42 & 13.43 & 83.36& 24.20 & 25.15 & 0.06 & 30.96 & 30.08 & 41.78
    \\ \midrule
    MTri~\citep{Zhang2018ICASSP} ($C_1$) (w/ ent) & 64.88 & 19.33 & 61.55 & 12.76 & 20.81 & 30.61 & \textbf{42.13} & 14.69 & 75.2 & 12.17 & 60.8 & \textbf{64.45} & \textbf{29.6} & 82.1 & 25.61 & 32.41 & 5.29 & 32.92 & 27.09 & 37.6 \\ 
    MTri~\citep{Zhang2018ICASSP} ($C_2$) (w/ ent) & 62.1 & 19.64 & 59.0 & 15.18 & 20.87 & 30.43 & 41.99 & 14.55 & 75.41 & 12.2 & 60.67 & 64.35 & 29.47 & 82.18 & 25.74 & 32.46 & 5.34 & 33.04 & 27.04 & 37.46 \\ 
    MTri~\citep{Zhang2018ICASSP} ($C_3$) (w/ ent) & 58.11 & 19.18 & 55.65 & 16.78 & 21.13 & 30.39 & 41.91 & 13.93 & 75.84 & 12.14 & 58.99 & 64.1 & 29.0 & 82.95 & 25.99 & \textbf{32.51} & \textbf{5.61} & \textbf{33.75} & 27.46 & 37.13\\ 
    \midrule
    Ours ($C_1$) (w/o ent) & \blueul{82.9} & 34.06 & 74.9 & 25.74 & 15.76 & 33.8 & 33.6 & 17.09 & 84.94 & 34.37 & 74.21 & 60.81 & 14.65 & 84.73 & 23.86 & 26.31 & \blueul{0.64} & 22.14 & 32.0 & 40.87 \\        
    Ours ($C_2$) (w/o ent) & 78.24 & 31.48 & 71.71 & 26.37 & 19.18 & \blueul{36.22}& 32.49 & 25.61 & 85.1 & 31.41 & \blueul{\textbf{84.28}} & 60.28 & \blueul{18.06} & 84.79 & 26.16 & 29.64 & 0.28 & 23.29 & 32.9 & 41.97 \\    
    Ours ($C_3$) (w/o ent) & 81.91 & 30.15 & \blueul{77.22} & 26.38 & 15.0 & 34.63 & 31.53 & \blueul{27.42} & 83.75 & 35.18 & 81.37 & 61.71 & 17.32 & \blueul{\textbf{85.07}} & \blueul{\textbf{26.5}} & 29.86 & 0.2 & 21.36 & 33.43 & 42.10 \\
    Ours ($C_m$) (w/o ent) & 82.78 & \blueul{35.74} & 75.81 & \blueul{26.83} & \blueul{19.89} & 34.96 & \blueul{34.47} & 25.60 & \blueul{\textbf{85.23}} & \blueul{\textbf{35.35}} & 79.75 & \blueul{62.02} & 14.82 & 84.68 & 25.30 & \blueul{32.05} & 0.03 & \blueul{27.48} & \blueul{41.23} & \blueul{43.39} \\ 
    Ours ($C_1$) (w/ ent) & 83.24 & 33.4 & 78.04 & 27.48 & 18.37 & 33.26 & 35.34 & 22.34 & \redul{83.87} & 27.47 & \redul{82.2} & 62.7 & 28.26 & 80.76 & 20.49 & 15.41 & 0.22 & 27.12 & 37.78 & 41.99 \\
    Ours ($C_2$) (w/ ent) & 84.67 & 33.64 & 80.30 & 27.33 & 19.37 & \redul{35.95} & 33.10 & 27.49 & 83.84 & 30.29 & 81.53 & 61.98 & 26.54 & 81.50 & 21.48 & 20.18& 0.03 & 29.05 & 40.57& 43.10 \\
    Ours ($C_3$) (w/ ent) & \redul{\textbf{87.22}} & \redul{\textbf{36.57}} & 81.26& 28.65 & 17.82& 35.55& 32.58 & 29.11 & 83.46 & 30.39 & 77.06& 62.48& 28.78& 81.49& 22.75 & 22.85 & 0.06 & 29.85 & 35.14 & 43.32 \\
    Ours ($C_m$) (w/ ent) & 85.29 & 35.57 & \redul{81.69} & \redul{29.93} & \redul{\textbf{20.24}} & 35.53 & \redul{36.63} & \redul{\textbf{35.94}} & 83.24 & \redul{28.1} & 81.75 & \redul{63.75} & \redul{29.18} & \redul{81.8} & \redul{23.44} & \redul{24.58} & \redul{4.67} & \redul{31.0} & \redul{\textbf{47.26}} & \redul{\textbf{45.24}} \\ \bottomrule    
    \end{tabular}
\end{adjustbox}

\end{table*}

\begin{equation}
\label{eq:stage2}
\begin{split}
  &\mathcal{L}_{stage2} =\mathcal{L}_{seg}(C_{k}(E(X^{T_k}_{\mathcal{S}})), Y_\mathcal{S}) +\lambda_{adv}\mathcal{L}_{adv}(X^{T_c}_\mathcal{S}, X_\mathcal{T}) \\
  &+ \lambda_{ent}\mathcal{L}_{ent}(C_{k}(E(X_{\mathcal{T}}))) + \mathcal{L}_{seg}(C_k(E(X_{\mathcal{T}})), \hat{Y}^{(i-1)}_\mathcal{T})
\end{split}
\end{equation}

$\forall k \in \{1, 2, 3\}$. Second, the meta-learner $C_m$ is retrained on the predictions of the three updated classifiers on the target images along with the pseudo-labels $\hat{Y}^{(i-1)}_\mathcal{T}$:

\begin{equation}
\label{eq:arbiter_optimization_ret}
\argmin_{\mathrm{w}_1, \mathrm{w}_2, \mathrm{w}_3} \mathcal{L}_{seg}(C_m(X_{\mathcal{T}}, X_{\mathcal{T}}, X_{\mathcal{T}}), \hat{Y}^{(i-1)}_\mathcal{T})
\end{equation}

And finally, with the updated weight vectors $\mathrm{w}_{k}$, the meta-learner is able to generate new pseudo-labels for the target images $\hat{Y}^{(i)}_\mathcal{T}$ to be used for the next rounds. The number of SSL rounds will be dictated by $C_m$, specifically until the performance gap between $C_m$ and the three classifiers $C_k$ is no longer significant. The entire training procedure is summarized in Algorithm~\ref{alg:training}.

\normalem 
\begin{algorithm}[htb]
\label{alg:training}
\DontPrintSemicolon
\SetKwInOut{Input}{Input}
\SetKwInOut{Output}{Output}
\SetAlgoLined
\Input{$(X_\mathcal{S}, Y_\mathcal{S})$, $(X_\mathcal{T}, Y_\mathcal{T} = \varnothing)$    }
\Output{$E$, $C_1$, $C_2$, $C_3$ and $C_m$}
obtain $X^{T_1}_{\mathcal{S}}$, $X^{T_2}_{\mathcal{S}}$ and $X^{T_3}_{\mathcal{S}}$ from $T$\tcp*[r]{stage 1} 
 train $E$, $C_1$, $C_2$, $C_3$ and $D$ with Eq.~(\ref{eq:stage1})\;
train $C_m$ with Eq.~(\ref{eq:arbiter_optimization}) using $(X_\mathcal{S}, Y_\mathcal{S})$\;
generate $\hat{Y}_\mathcal{T}^{(0)}$ from $C_m$ using $X_\mathcal{T}$\tcp*[r]{stage 2} 
 \For{$i\leftarrow 1$ \KwTo $\text{number of rounds}$}{
     train $E$, $C_1$, $C_2$, $C_3$ and $D$ with Eq.~(\ref{eq:stage2})\;
     retrain $C_m$ with Eq.~(\ref{eq:arbiter_optimization_ret}) using $(X_\mathcal{T}, \hat{Y}_\mathcal{T}^{(i-1)})$\;     
     generate $\hat{Y}_\mathcal{T}^{(i)}$ from $C_m$ using $X_\mathcal{T}$\;
 }
 \caption{Training process of our method}
\end{algorithm}
\ULforem 

\section{Experiments}
\label{sec:experiments}

Since the goal in UDA involves adapting a source annotated domain to an unlabeled target domain, we use the challenging synthetic-to-real UDA benchmarks for semantic segmentation to prove our main hypotheses as well as compare our approach with state-of-the-art methods. In this set-up, models are trained jointly with fully-annotated synthetic data as well as unlabeled real-world data, both considered source and target domain respectively. The models are then validated on an unseen portion of the target domain, measuring their mean intersection-over-union (mIoU)~\citep{Everingham2014miou}.

Considering that our approach aims to improve on the discrepancy loss proposed in~\citep{Zhang2018ICASSP} by training each individual classifier with a different set of features and using a meta-learning layer is to ensemble the classifiers' outputs, the first experiment focuses on analyzing the difference between Multi-task tri-training (MTri)~\citep{Zhang2018ICASSP} and our method (see Fig.~\ref{fig:eyecatch}) during the first training stage. This comparison is complemented with a (vanilla) single encoder-decoder (SED) network architecture, often used in the literature for UDA~\citep{Li2019BDLCVPR, yang2020fda, yang2020labeldriven, Zou2018eccv}. The goal of this experiment is to study how the meta-learner can balance the outputs of the classifiers with respect to the classes, and how this compares to the aforementioned existing UDA architectures. Additionally, we also show the distribution of the weights for the meta-learner over all the classes and analyze the influence of entropy minimization.

The second experiment compares our proposed UDA approach with current state-of-the-art methods, putting our model into context with approaches that rely on a combination of image translation, feature matching, entropy minimization and SSL.

Finally, although most UDA methods focus only on analyzing their approach on the standard benchmarks of adapting synthetic-to-real domains, they tend to neglect the generalization capacity of the resulting model. For this reason, we designed a third experiment to study how well our model can generalize to a completely unseen dataset that was not used during the UDA training process. The protocol for this experiment consists of first adapting GTA V~\citep{Richter2016ECCV} to Cityscapes~\citep{Cordts2016Cityscapes}, and then evaluating the resulting model on WildDash~\citep{Zendel_2018_ECCV}. To compare with other state-of-the-art methods, we select those whose code is publicly available and provide an evaluation script, and proceed to 1) reproduce their result on the proposed synthetic-to-real benchmark and 2) evaluate the model on WildDash.

\subsection{Datasets}

\textbf{Target dataset.} Cityscapes~\citep{Cordts2016Cityscapes} is a large-scale and real urban scene semantic segmentation dataset that provides $5000$ finely annotated images split into three sets: train ($2975$), validation ($500$) and test ($1525$). These sets are pixel-wise labeled, with a resolution of $1024 \times 2048$  pixels. The number of classes is $34$ but only $19$ are officially considered in the evaluation protocol.

\textbf{Source datasets.} GTA V~\citep{Richter2016ECCV} is a synthetic dataset that contains $24966$ labeled frames taken from a realistic open-world computer game called Grand Theft Auto V (GTA V). The resolution of the images is $1052 \times 1914$ pixels and most of the frames are vehicle-egocentric. All the classes are compatible with the $19$ official classes of Cityscapes. SYNTHIA~\citep{synthia} is a synthetic dataset consisting of driving scenes rendered from a virtual city. We use the SYNTHIA-RAND-CITYSCAPES subset as source set, which contains $9400$ $1280 \times 760$ images for training and $16$ common classes with Cityscapes, and we evaluate the resulting model on these 16 classes.

\textbf{Unseen dataset.} WildDash~\citep{Zendel_2018_ECCV} is a real-world dataset containing $4256$ finely annotated images in a pixel-wise manner, created with the purpose of testing the robustness of models under different driving scenarios (e.g. rain, road coverage, darkness, overexposure). These images have a resolution of $1920 \times 1080$ pixels and the labels are fully compatible with Cityscapes.

\begin{figure}[t]
 \centering 
\includegraphics[width=0.95\linewidth]{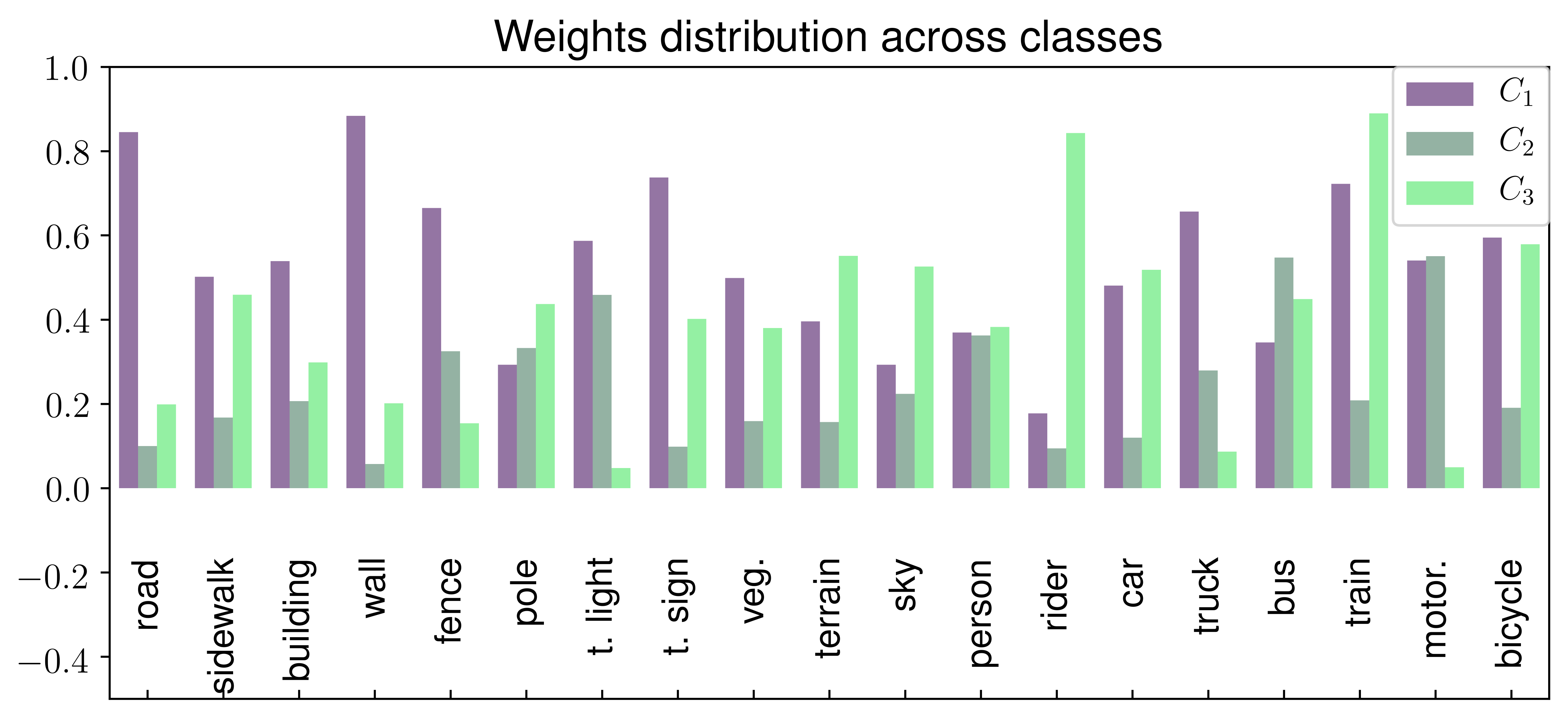}
 \caption{\textit{weights after optimizing Eq.~(\ref{eq:arbiter}) for the adaption GTA V $\xrightarrow{}$ Cityscapes.} Before starting SSL, $C_1$ and $C_3$ are the most dominant predictors on the output space of $C_m$.} 
 \label{fig:afteradapt}
\end{figure}

\subsection{Implementation details}
\label{sec:impdetails}

\begin{table*}[!ht]
\centering
\caption{\label{tab:abstudy2}Adapting from GTA V to Cityscapes. S1 and S2 indicate the training stage, while R1 and R2 denote the first and second round of SSL, respectively.}    
\begin{adjustbox}{width=0.95\textwidth}
    \begin{tabular}{l | ccccccccccccccccccc| c }
    \toprule
    \multicolumn{21}{c}{\textbf{GTA V} $\xrightarrow{}\textbf{Cityscapes}$} \\
    \midrule
     Method & \rotatebox{60}{road} & \rotatebox{60}{side.} & \rotatebox{60}{buil.} & \rotatebox{60}{wall} & \rotatebox{60}{fence} &  \rotatebox{60}{pole} & \rotatebox{60}{light} & \rotatebox{60}{sign} & \rotatebox{60}{veget.} & \rotatebox{60}{terr.} & \rotatebox{60}{sky} & \rotatebox{60}{person} & \rotatebox{60}{rider} & \rotatebox{60}{car} & \rotatebox{60}{truck} & \rotatebox{60}{bus} & \rotatebox{60}{train} & \rotatebox{60}{motor} & \rotatebox{60}{bike} & \rotatebox{60}{\textbf{mIoU}}\\ \midrule
    DCAN~\citep{Wu2018acm}& 85.0& 30.8& 81.3& 25.8& 21.2& 22.2& 25.4& 26.6& 83.4& 36.7& 76.2& 58.9& 24.9& 80.7 &29.5& 42.9& 2.5 &26.9 &11.6& 41.7 \\ 
    DLOW~\citep{Gong_2019}& 87.1& 33.5& 80.5& 24.5& 13.2& 29.8& 29.5& 26.6& 82.6& 26.7 &81.8& 55.9& 25.3 &78.0 &33.5 &38.7 &0.0 &22.9 &34.5& 42.3 \\ 
    CLAN~\citep{Luo2018cvpr} & 87.0 & 27.1 & 79.6 & 27.3 & 23.3 & 28.3 & 35.5 & 24.2 & 83.6 & 27.4 & 74.2 & 58.6 & 28.0 & 76.2 & 33.1 & 36.7 & 6.7 & 31.9 & 31.4 & 43.2 \\ 
    ABStruct~\citep{Chang_2019} & 91.5& 47.5& 82.5& 31.3& 25.6& 33.0& 33.7& 25.8& 82.7 &28.8& 82.7& 62.4& 30.8& 85.2& 27.7& 34.5& 6.4& 25.2& 24.4& 45.4 \\ 
    ADVENT~\citep{TH2018ADVENTCVPR} & 89.4 &33.1& 81.0& 26.6& 26.8& 27.2& 33.5& 24.7& 83.9& 36.7& 78.8& 58.7& 30.5& 84.8& 38.5& 44.5& 1.7& 31.6& 32.4& 45.5 \\ 
    BDL~\citep{Li2019BDLCVPR} & 91.0 &44.7& 84.2& 34.6& 27.6& 30.2& 36.0& 36.0& 85.0& \textbf{43.6}& 83.0& 58.6& 31.6& 83.3& 35.3& 49.7& 3.3& 28.8& 35.6& 48.5 \\ 
    FDA-MBT~\citep{yang2020fda} & 92.5& \textbf{53.3}& 82.4 &26.5& 27.6& 36.4& 40.6& 38.9& 82.3& 39.8& 78.0& 62.6& \textbf{34.4}& 84.9& 34.1& \textbf{53.1}& 16.9& 27.7& 46.4& 50.45 \\ 
    PCEDA~\citep{Yang_2020} & 91.0 &49.2& 85.6& \textbf{37.2}& \textbf{29.7}& 33.7& 38.1& 39.2& 85.4& 35.4& 85.1& 61.1 &32.8& 84.1& \textbf{45.6}& 46.9& 0.0& 34.2& 44.5& 50.5 \\ \midrule
    Ours S1 ($C_3$) & 87.22 & 36.57 & 81.26& 28.65 & 17.82& 35.55& 32.58 & 29.11 & 83.46 & 30.39 & 77.06& 62.48& 28.78& 81.49& 22.75 & 22.85 & 0.06 & 29.85 & 35.14 & 43.32 \\
    Ours S1 ($C_m$) & 85.29 & 35.57 & 81.69 & 29.93 & 20.24 & 35.53 & 36.63 & 35.94 & 83.24 & 28.1 & 81.75 & 63.75 & 29.18 & 81.8 & 23.44 & 24.58 & 4.67 & 31.0 & 47.26  & 45.24 \\
    Ours S2-R1 ($C_3$) & 90.6 & 46.94 & 84.06 & 31.9 & 23.88 & 37.53 & 34.81 & 34.37 & 85.69 & 36.02 & 84.32 & 66.53 & 29.41 & 85.46 & 27.77 & 32.48 & 7.15 & \textbf{36.05} & 54.96 & 48.94 \\
    Ours S2-R1 ($C_m$) & 90.81 & 47.85 & 85.01 & 32.08 & 24.55 & 37.73 & 38.15 & 42.13 & 85.37 & 34.32 & 84.97 & 66.51 & 28.0 & 84.51 & 27.0 & 25.14 & 15.23 & 35.03 & 56.02 & 49.50 \\
    Ours S2-R2 ($C_2$) & 92.27 & 51.59 & 86.19 & 35.28 & 26.84 & 36.73 & 35.68 & 42.4 & \textbf{86.84} & 37.3 & \textbf{85.49} & 66.9 & 27.6 & 85.75 & 32.26 & 32.85 & \textbf{20.59} & 33.89 & \textbf{58.1} & 51.29 \\
    Ours S2-R2 ($C_m$) & \textbf{92.59} & 53.05 & \textbf{86.31} & 34.2 & 27.17 & \textbf{39.13} & \textbf{41.0} & \textbf{44.8} & 86.1 & 34.32 & 84.69 & \textbf{67.23} & 29.77 & \textbf{85.78} & 32.73 & 29.9 & 20.12 & 35.55 & 57.05 & \textbf{51.66} \\
    \bottomrule    
    \end{tabular}
\end{adjustbox}
\caption{\label{tab:abstudy3}Adapting from SYNTHIA to Cityscapes. Total mIoU values with * are reported only on 13 subclasses (excluding \textit{wall}, \textit{fence} and \textit{pole}). Our method achieves the best performance over all the 16 classes.}
\begin{adjustbox}{width=0.95\textwidth}
    \begin{tabular}{l | cccccccccccccccc| c }
    \toprule
    \multicolumn{18}{c}{\textbf{SYNTHIA} $\xrightarrow{}\textbf{Cityscapes}$} \\
    \midrule
     Method & \rotatebox{60}{road} & \rotatebox{60}{side.} & \rotatebox{60}{buil.} & \rotatebox{60}{wall*} & \rotatebox{60}{fence*} &  \rotatebox{60}{pole*} & \rotatebox{60}{light} & \rotatebox{60}{sign} & \rotatebox{60}{veget.} & \rotatebox{60}{sky} & \rotatebox{60}{person} & \rotatebox{60}{rider} & \rotatebox{60}{car} & \rotatebox{60}{bus} & \rotatebox{60}{motor} & \rotatebox{60}{bike} & \rotatebox{60}{\textbf{mIoU}} \\ \midrule 
    AdaptPatch~\citep{Tsai2019ICCV} & 82.4 & 38.0 & 78.6 & 8.7 & 0.6 & 26.0 & 3.9 & 11.1 & 75.5 & 84.6 & 53.5 & 21.6 & 71.4 & 32.6 & 19.3 &   31.7 & 40.0 \\
    AdaptSegNet~\citep{Tsai2018CVPR} & 79.2 & 37.2 & 78.8 & - & - & - & 9.9 & 10.5& 78.2 & 80.5 & 53.5 & 19.6 & 67.0 & 29.5 & 21.6 & 31.3 & 45.9* \\ 
    BDL~\citep{Li2019BDLCVPR} & \textbf{86.0} & \textbf{46.7} & 80.3 & - & - & - & 14.1 & 11.6 & 79.2 & 81.3 & 54.1 & 27.9 & 73.7 & \textbf{42.2} & 25.7 & 45.3& 51.4* \\
    CLAN~\citep{Luo2018cvpr} & 81.3 & 37.0 & 80.1 & - & - & - & 16.1 & 13.7 & 78.2 & 81.5 & 53.4 & 21.2 & 73.0 & 32.9 & 22.6 & 30.7 & 47.8* \\
    FDA-MBT~\citep{yang2020fda} & 79.3 & 35.0 & 73.2 & - & - & - & 19.9 & 24.0 & 61.7 & 82.6 & 61.4 & \textbf{31.1} & 83.9 & 40.8 & \textbf{38.4} & \textbf{51.1} & 52.5* \\
    PCEDA~\citep{Yang_2020} & 85.9 & 44.6 & \textbf{80.8} & \textbf{9.0} & \textbf{0.8} & 32.1 & 24.8 & 23.1 & 79.5 & 83.1 & 57.2 & 29.3 & 73.5 & 34.8 & 32.4 & 48.2 & 46.2 \\ 
    Ours ($C_m$) & 81.90 & 41.88 & 78.21 & 3.38 & 0.02 & \textbf{44.76} & \textbf{24.82} & \textbf{27.17} & \textbf{86.59} & \textbf{85.18} & \textbf{68.74} & 30.55 & \textbf{84.65} & 24.42 & 20.12 & 40.77 & \textbf{46.45} \\
    \bottomrule    
    \end{tabular}
\end{adjustbox}

\end{table*}

\begin{table}[t]
\centering
\caption{\label{tab:cainits}Number of classes where each classifier outperforms the others on  GTA V $\xrightarrow{}$ Cityscapes. Although there is a clear dominance of $C_3$ before starting SSL, this trend tends to wear off as the self supervision process unfolds.}
\begin{adjustbox}{width=0.75\columnwidth}
\begin{tabular}{c | c | c | c}
    classifier & stage 1 & SSL: round 1 & SSL: round 2\\ \hline
    $C_1$ & 5  & 6 & 4 \\
    $C_2$ & 4  & 7 & 7 \\    
    $C_3$ & 10 & 6 & 8 
\end{tabular}
\end{adjustbox}
\end{table}

\begin{table}[t]
\centering
\caption{\label{tab:tinfluencessl}Influence of $T$ on GTA V $\xrightarrow{}$ Cityscapes during first round of SSL. Using $T$ during SSL by feeding a transformation to its corresponding classifier provokes a slight performance drop in the mIoU.}
\begin{adjustbox}{width=0.75\columnwidth}
\begin{tabular}{c | c | c | c}
    classifier & stage 1 & SSL without $T$ & SSL with $T$\\ \hline
    $C_1$ & 41.99  & 48.87 & 47.9 \\
    $C_2$ & 43.10  & 48.91 & 47.8 \\    
    $C_3$ & 43.32 & 48.94 & 48.0 
\end{tabular}
\end{adjustbox}

\end{table}

\textbf{Image translation module and discriminator}. We have chosen CycleGAN~\citep{Zhu2017Unpaired} as our image translation model $T$, as it is a commonly used approach that can be trained efficiently. Its architecture provides two image translations: the translation from the source to the target domain $X^t_\mathcal{S}$ and the reconstruction back to the source set $X^r_\mathcal{S}$. These two translations are combined with the original source images to obtain three different representation $X^{T_1}_{\mathcal{S}} = X_{\mathcal{S}}$, $X^{T_2}_{\mathcal{S}} = X^r_{\mathcal{S}}$ and $X^{T_3}_{\mathcal{S}} = X^t_{\mathcal{S}}$. Regarding the discriminator $D$, feature alignment between target-like and target images is frequently done when using CycleGAN's as image translation module~\citep{Li2019BDLCVPR, Hoffman18ICML}, and therefore $X^{T_c}_{\mathcal{S}}$ turns into $X^{T_3}_{\mathcal{S}}$ in Eq.~(\ref{eq:advtraining}). We note that potentially better performing image translations approaches exist but in our work we opt for the commonly used approach CycleGAN. 

\textbf{Training protocols for MTri and SED}. We have respected the protocol for MTri as reported in~\citep{Zhang2018ICASSP}, i.e. minimizing a cross-entropy loss for semantic segmentation combined with a cosine distance loss for the discrepancy between $C_1$ and $C_2$. As for the single encoder-decoder (SED) approach, all our losses were implemented using one encoder and one classifier, while using all three available transformations. In essence, the SED approach is similar to our approach but does not use the ensemble approach with the three classifiers.

\textbf{Hardware and network architecture}. In our experiments, we have implemented our method using Tensorflow~\citep{tensorflow} and trained our model using a single NVIDIA TITAN RTX with 24 GB memory. Regarding the segmentation network, we have chosen ResNet101~\citep{Kaiming2015CVPR} pretrained on ImageNet~\citep{imagenet2009cvpr} as feature extractor for $E$. When it comes to the decoders $C_k, k \in \{1, 2, 3\}$, DeepLab-v2~\citep{Deeplabv2} framework was used. For the network $D$ we adopt a similar structure than~\citep{Radford2015ungan}, which consists of $5$ convolution layers with kernel size of $4\times4$, stride $2$ and channel numbers \{4, 8, 16, 32, 1\}. Each layer uses Leaky-ReLU~\citep{Maas2013RectifierNI} as activation function with a slope of $0.2$, except the last layer that has no activation. Throughout the training process, we use SGD~\citep{SGD} as optimizer with momentum of $0.9$, encoder and decoders follow a poly learning rate policy, where the initial learning rate is set to $2.5e^{-4}$. The discriminator is also optimized with SGD but with a fixed learning rate of $1e^{-5}$. As for the entropy loss, we chose $\lambda_{ent}=0.005$ and $\eta=2.0$ for all experiments. During the first stage the network is trained for $150k$ iterations. Then we perform SSL until convergence is reached on each round. We use a crop size of $512 \times 1024$ during training, and we evaluate on full resolution $1024 \times 2048$ images from Cityscapes validation split.

\subsection{Our approach vs MTri vs single encoder-decoder (SED)}
\label{sec:mtrivssingle}

We can infer from Tab.~\ref{tab:abstudy1} that our method outperforms both SED and MTri, and the difference is even more noticeable when comparing against the meta-learner. $C_m$ performs better than the individual classifiers over $14$ classes if we make the comparison with entropy minimization and over $11$ classes without entropy minimization. It is also important to mention that there is a considerable correlation between the weights of the meta-learner that are depicted in Fig.~\ref{fig:afteradapt} and the performance of the meta-learner. If we analyze the classes in Tab.~\ref{tab:abstudy1} where $C_m$ outperforms the three classifiers (for instance: \textit{fence}, \textit{traffic light}, \textit{rider}, \textit{bus}, \textit{bike}), we can see also a pattern in Fig.~\ref{fig:afteradapt} where the meta-learner tends to amplify the contribution of the two best performing classifiers whilst penalizing the one with the lowest mIoU. For some other classes such as \textit{traffic sign} and \textit{truck}, it prefers to perform a mixture of weak classifiers, resulting in a final prediction that outperforms the strongest member of the ensemble.

\begin{figure}[t]
 \centering 
 \includegraphics[width=0.95\columnwidth]{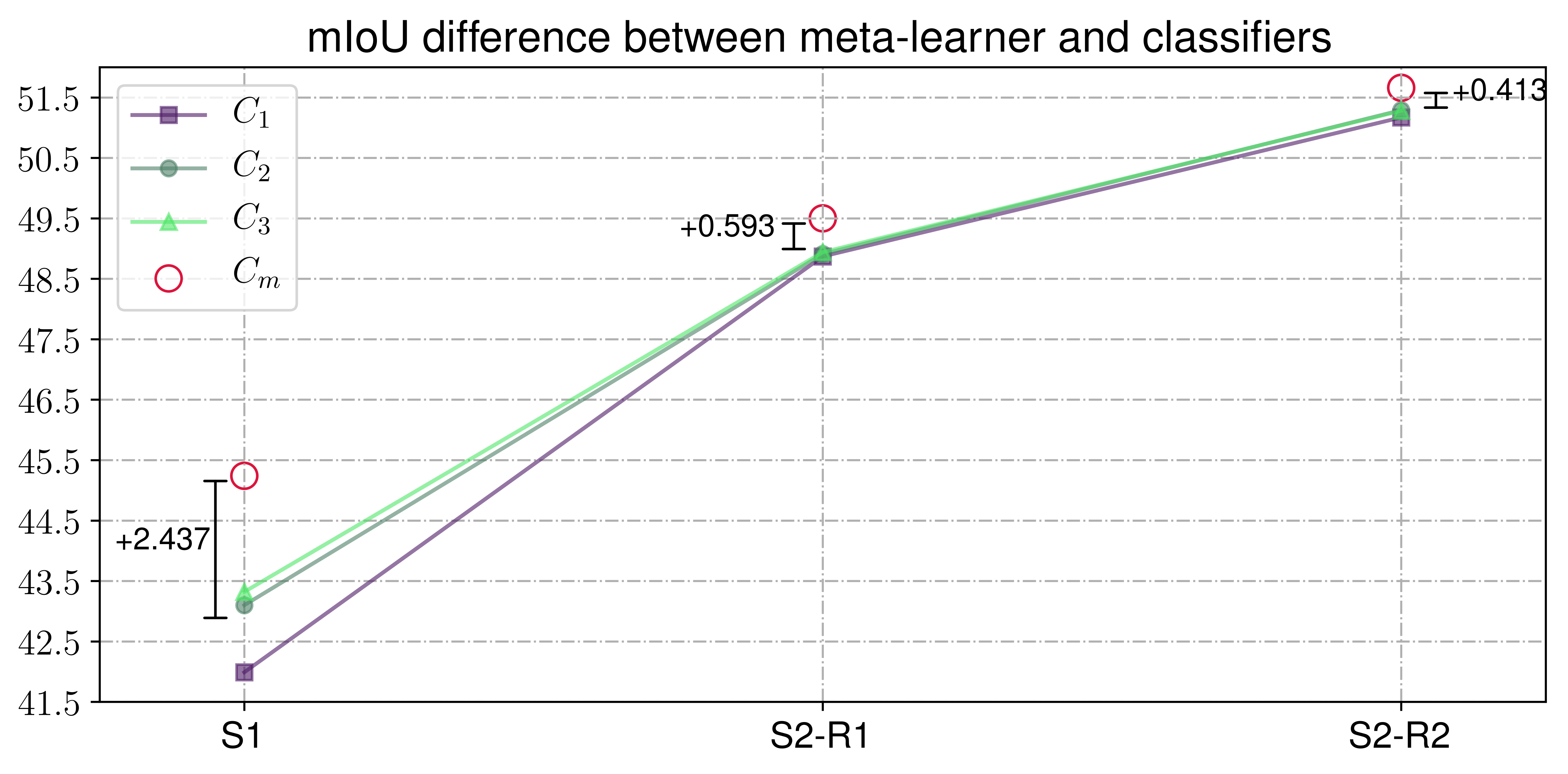}
 \caption{\textit{Effect of SSL for $C_m$.} Retraining the meta-learner after each round of SSL makes sure that it keeps outperforming the other classifiers with a decreasing margin.} 
 \label{fig:miou_gaps}
\end{figure}

\begin{table}[t]

\centering
\begin{adjustbox}{width=0.75\columnwidth}
\begin{tabular}{c | c | c | c}
    Method & \# encoders & \# classifiers & mIoU \\ \hline
    ADVENT~\citep{TH2018ADVENTCVPR} & 1 & 1 & 25.9 \\
    BDL~\citep{Li2019BDLCVPR} & 1 & 1 & 26.57 \\    
    FDA-MBT~\citep{yang2020fda} & 3 & 3 & 31.07 \\
    Ours & 1 & 3 & \textbf{31.2}
\end{tabular}
\end{adjustbox}
\caption{\label{tab:generalizationtest}Generalization test on WildDash after adapting  GTA V $\xrightarrow{}$ Cityscapes.}
\end{table}

\subsection{Our approach vs state-of-the-art methods}
\label{sec:mtrivssota}

\begin{figure*}
\setlength\tabcolsep{0.5pt}
\settowidth\rotheadsize{Image}
\resizebox{\textwidth}{0.09\textheight}{
\begin{tabularx}{\linewidth}{l XXXXX }
\rothead{\centering Image} &   \includegraphics[width=\hsize,valign=m]{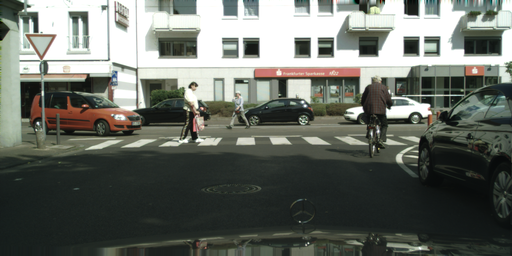}
                &   \includegraphics[width=\hsize,valign=m]{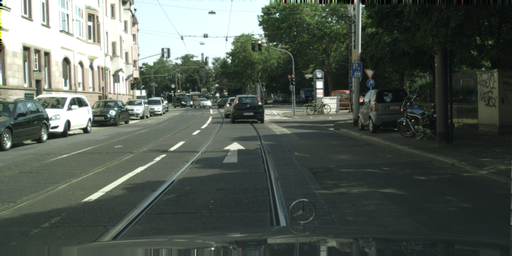}
                &   \includegraphics[width=\hsize,valign=m]{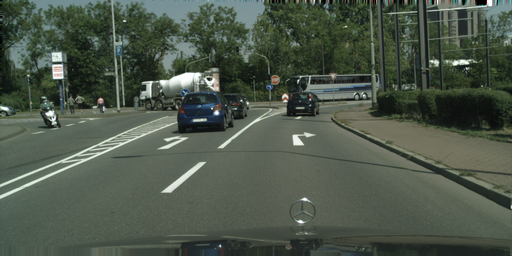}
                &   \includegraphics[width=\hsize,valign=m]{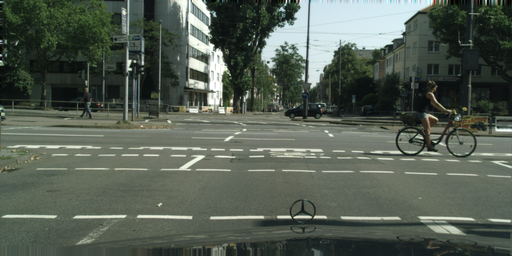}                
                &   \includegraphics[width=\hsize,valign=m]{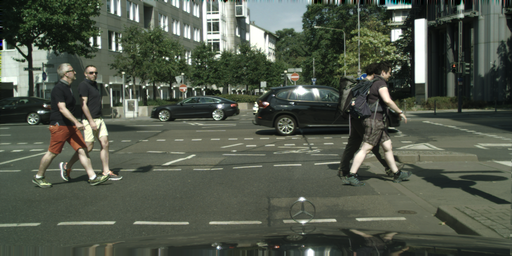}\\ \addlinespace[0.5pt]
\rothead{\centering GT} &   \includegraphics[width=\hsize,valign=m]{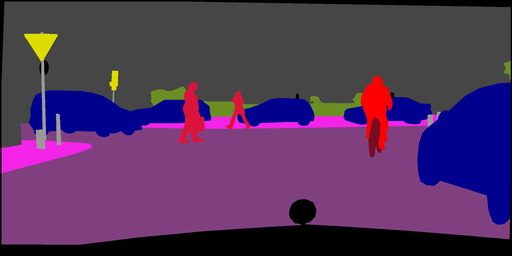}
                &   \includegraphics[width=\hsize,valign=m]{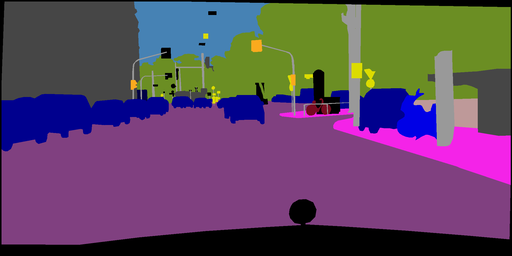}
                &   \includegraphics[width=\hsize,valign=m]{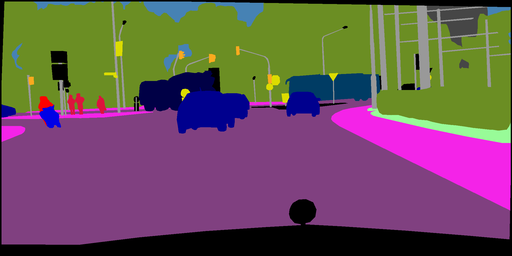}
                &   \includegraphics[width=\hsize,valign=m]{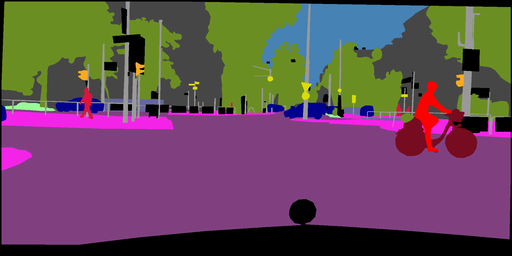}                
                &   \includegraphics[width=\hsize,valign=m]{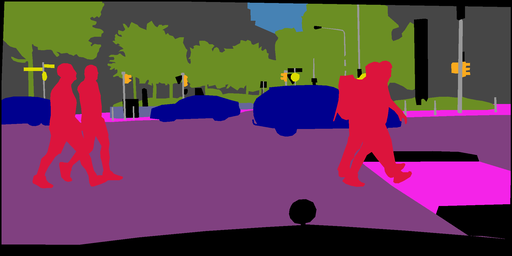}\\ \addlinespace[0.5pt]
\rothead{\centering $C_m$} &   \includegraphics[width=\hsize,valign=m]{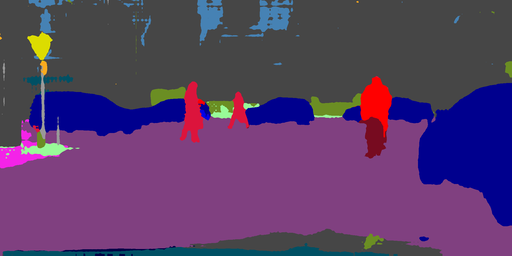}
                &   \includegraphics[width=\hsize,valign=m]{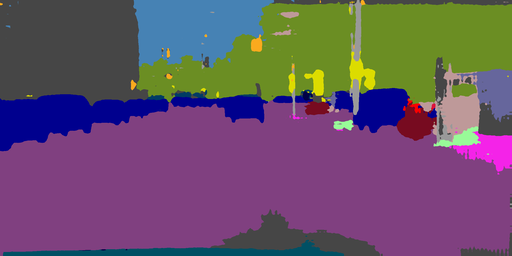}
                &   \includegraphics[width=\hsize,valign=m]{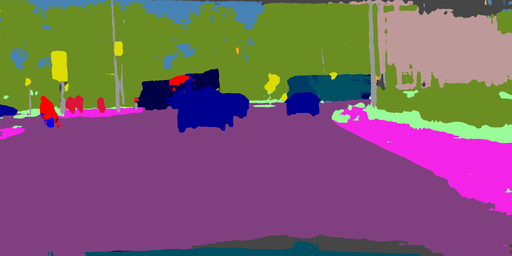}
                &   \includegraphics[width=\hsize,valign=m]{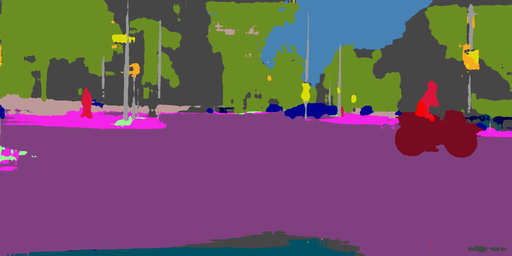}                
                &   \includegraphics[width=\hsize,valign=m]{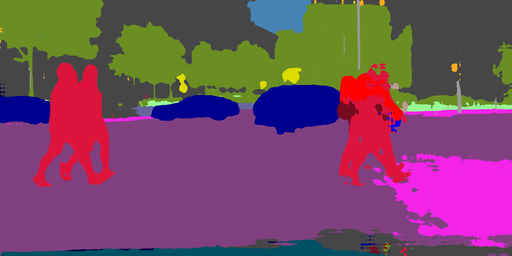}
\end{tabularx}
}
\caption{\textit{Qualitative comparison from GTA V to Cityscapes}. The meta-learner rebalances the predictions from $C_1$, $C_2$ and $C_3$ to achieve a smoother output over all the classes, where less predominant classes such as \textit{t. sign} and \textit{pole} have more presence on the pixel space of $C_m$.}
\label{tab:gtav2cityquali}
\end{figure*}

The quantitative results for the adaption GTA V $\xrightarrow{}$ Cityscapes can be seen in Tab.~\ref{tab:abstudy2}. When comparing it with state-of-the-art methods, we can see that our approach outperforms PCEDA~\citep{Yang_2020},  FDA-MBT~\citep{yang2020fda} and BDL~\citep{Li2019BDLCVPR}, three recent methods that use a combination of strategies. Qualitative results in Fig.~\ref{tab:gtav2cityquali} shows satisfactory results on the output space, leading to consistently clean predictions.

As for SYNTHIA $\xrightarrow{}$ Cityscapes, the mIoU values are shown in Tab.~\ref{tab:abstudy3}. We achieved competitive results over all 16 classes with respect to state-of-the-art methods such as PCEDA~\citep{Yang_2020} and AdaptPatch~\citep{Tsai2019ICCV}, dominating on some difficult classes such as \textit{pole}, \textit{traffic light} and \textit{traffic sign}.

If we analyze the improvements during SSL, we see that the meta-learner consistently scores better than the individual three classifiers, although its gain diminishes with each round of SSL (see Fig.~\ref{fig:miou_gaps}). This can be attributed to the fact that all three classifiers are being optimized with the same images, and thus the model is losing the capability to keep diversity among the predictors. The results from Tab.~\ref{tab:abstudy2} and Tab.~\ref{tab:cainits} also show that the dominance of the members of the ensemble can alternate since $C_3$ is the best predictor after the first round (R1) and $C_2$ takes over after the second one (R2). This suggests that some predictors can learn more than others, even if they share the same input images, showing that all of them are equally important. This can be better appreciated in Tab.~\ref{tab:cainits} where $C_2$ stands out after R1 and remains close to $C_3$ after R2 when analyzing the mIoU per class during SSL.

Using the image translation module $T$ during the second stage by transforming the target set into the closest transformation possible to each classifier, i.e., transforming the target images to the source domain for the first two classifiers while keeping them unaltered for the third one, leads to slightly worse performance (see Tab.~\ref{tab:tinfluencessl}). This can be attributed to the fact that, since SSL aims to close the gap for the target distribution, it is needed to keep the inputs as similar as possible to those that the algorithm would receive during inference.

\subsection{Generalization capacity of our approach}
\label{sec:genexperiment}

The results of the proposed generalization test in Tab.~\ref{tab:generalizationtest} shows different UDA methods along with their arrangement for the semantic segmentation network and the corresponding mIoU performance on WildDash, after adapting GTA V to Cityscapes. ADVENT~\citep{TH2018ADVENTCVPR} is a UDA approach that does not leverage any image translation strategy, while BDL~\citep{Li2019BDLCVPR} and FDA-MBT~\citep{yang2020fda} make use of one and three image translations respectively. If we consider the amount of encoders and classifiers, we can notice that using a single encoder-decoder gives a quite limiting generalization performance, although BDL outperforms ADVENT. This slightly better performance of BDL can be attributed to the usage of one image translation (transforming the source to the target with CycleGAN) to increase the robustness of the model.

FDA-MBT uses three image representations, mapping the source annotated images to the target using three different parameters for the image translation module, and performing UDA by training each encoder-decoder segmentation model with a specific representation. The reported performance of $31.07$ is the result of averaging the three trained models, and although it is close to ours, our semantic segmentation network takes up significantly fewer parameters to train (one encoder and three classifiers). This makes our approach attractive as it is a good trade-off between its number of parameters and performance. More importantly, this experiment shows the advantageous effect of using multiple image translations, as done in FDA-MBT and our approach, on the generalizability of the trained models.

\section{Conclusions}

In this work, we used challenging synthetic-to-real semantic segmentation UDA benchmarks to verify our main hypotheses that self-supervised learning for UDA can be improved by: 1) using multiple image translations instead of a discrepancy loss to encourage diversity of classifiers when generating the pseudo-labels, and 2) adding a meta-learner that utilizes the classifiers to improve the quality and robustness of the obtained pseudo-labels.

We have shown empirically in Section~\ref{sec:mtrivssingle} and Section~\ref{sec:mtrivssota} that the proposed approach, which combines the benefits of image translations, self-supervised learning, and ensemble learning, improves the model's accuracy for two standard UDA benchmarks when adapting synthetic to real data and also shows satisfactory generalization capacity as discussed in Section~\ref{sec:genexperiment}.

We can conclude from the results that increasing the input variability via different image translations induces the network to learn better domain agnostic representations in the feature extractor while keeping specificity on each classifier. Using ensemble learning to integrate the outputs of the classifiers is also beneficial as it allows the model to generate high-quality pseudo-labels and thereby improve the self-supervised learning process. 

We should remember that although we focused on the standard synthetic-to-real UDA benchmarks, it is also possible to extend this work to real-to-real applications where both source and target domains are real-world datasets. While this can represent a more realistic application of UDA, we consider that in this work we made a step in improving the generalization capability of deep learning models.

\bibliographystyle{model2-names}
\bibliography{references}

\end{document}